\documentclass{article}

\usepackage[final]{neurips_2024}

\usepackage[utf8]{inputenc} % allow utf-8 input
\usepackage[T1]{fontenc}    % use 8-bit T1 fonts
\usepackage{hyperref}       % hyperlinks
\usepackage{url}            % simple URL typesetting
\usepackage{booktabs}       % professional-quality tables
\usepackage{amsfonts}       % blackboard math symbols
\usepackage{nicefrac}       % compact symbols for 1/2, etc.
\usepackage{microtype}      % microtypography
\usepackage{xcolor}         % colors
\usepackage{multirow}
\usepackage{algorithm}
\usepackage{algpseudocode}
\usepackage{amsmath}
\usepackage{graphicx}  
\usepackage{svg}
\usepackage{caption}
\usepackage{subcaption}
\usepackage{bm}
\usepackage{makecell}

\usepackage[style=numeric]{biblatex} 
\title{DuoDiff: Accelerating Diffusion Models with a Dual-Backbone Approach}

\author{%
  Daniel Gallo Fernández\thanks{Alphabetical order. Equal contribution} \\
  University of Amsterdam \\\texttt{daniel.gallo.fernandez@student.uva.nl} \\
  \And
  Răzvan-Andrei Matişan$^*$ \\
  University of Amsterdam \\
  \texttt{razvan.matisan@student.uva.nl} \\
  \And
  Alejandro Monroy Muñoz$^*$ \\
  University of Amsterdam \\
  \texttt{alejandro.monroy.munoz@student.uva.nl} \\
  \And
  Ana-Maria Vasilcoiu$^*$ \\
  University of Amsterdam \\
  \texttt{ana-maria.vasilcoiu@student.uva.nl} \\
  \And
  Janusz Partyka \\
  University of Amsterdam \\
  \texttt{janusz.partyka@student.uva.nl} \\
  \And
  Tin Hadži Veljković \\
  University of Amsterdam \\
  \texttt{t.hadziveljkovic@uva.nl} \\
  \And
  Metod Jazbec \\
  University of Amsterdam \\
  \texttt{m.jazbec@uva.nl} \\
}

\begin{document}

\maketitle

\begin{abstract}
Diffusion models have achieved unprecedented performance in image generation, yet they suffer from slow inference due to their iterative sampling process. To address this, early-exiting has recently been proposed, where the depth of the denoising network is made adaptive based on the (estimated) difficulty of each sampling step. Here, we discover an interesting ``phase transition'' in the sampling process of current adaptive diffusion models: the denoising network consistently exits early during the initial sampling steps, until it suddenly switches to utilizing the full network. Based on this, we propose accelerating generation by employing a shallower denoising network in the initial sampling steps and a deeper network in the later steps. We demonstrate empirically that our dual-backbone approach, \emph{DuoDiff}, outperforms existing early-exit diffusion methods in both inference speed and generation quality. Importantly, DuoDiff is easy to implement and complementary to existing approaches for accelerating diffusion.  
\end{abstract}

\section{Introduction}

Diffusion models \cite{sohl2015deep} have recently demonstrated impressive performance in generative tasks across various modalities, including images \cite{ddpm, dhariwal2021diffusion}, videos \cite{ho2022imagen, ho2022video}, audio \cite{kong2020diffwave}, and molecules \cite{hoogeboom2022equivariant}. However, generating new samples with diffusion can be slow, as numerous sequential calls to the denoising network are required \cite{tomczak2022deep}. To improve sampling efficiency \cite{ulhaq2022efficient}, some of the most promising approaches focus on reducing the number of sampling steps (e.g., DDIM \cite{ddim} and distillation-based methods \cite{distil2, distil1}) or modifying the sampling space (e.g., latent diffusion \cite{rombach2022high}). 

Complementary to these efforts to accelerate diffusion, early-exiting \cite{teerapittayanon2016branchynet} has been proposed in AdaDiff \cite{adadiff}. Unlike the aforementioned static methods, AdaDiff is an adaptive approach in which the utilized depth of the denoising network can vary between sampling steps. Specifically, the difficulty of each sampling step $t$ (where $t$ decreases from the total number of steps $T$ to 0) is estimated by computing the uncertainty of the denoising network at each layer. If the uncertainty is low enough, the forward pass terminates at that layer (i.e., the model \emph{exits early}), thereby reducing computation for that step.

In this work, we leverage the adaptive nature of early-exit models to study the dynamics of the generative process in diffusion models. Interestingly, we find that early in the process (i.e., for large $t$), only a few layers of the denoising network are active, whereas later in the process (i.e., when $t$ approaches 0), the full network is utilized (Figure \ref{fig:ee-trends}). This suggests that the generation process in diffusion models begins with an easier phase, followed by a more challenging one. Motivated by these findings, we propose eliminating dynamic early-exit at every sampling step and instead introduce a (static) dual-backbone design, DuoDiff. DuoDiff consists of two denoising networks: a shallower one employed during the initial, easier phase of the generation process, and a deeper one used in the subsequent, more difficult stage (Figure \ref{fig:duodiff}). 

We experimentally demonstrate that DuoDiff outperforms existing early-exit diffusion models in both sampling latency and image generation quality across a range of standard datasets (e.g., ImageNet $256 \times 256$). Furthermore, compared to early-exit counterparts \cite{adadiff, moon2023early}, DuoDiff is better suited for batch inference, as it does not require per-sample computational paths. Additionally, we show that DuoDiff can be effectively combined with other popular efficiency-enhancing methods \cite{ddim,rombach2022high}.

\section{Background}

\textbf{Diffusion models} generate high-quality samples by progressively adding noise to data and learning to reverse this process. The \emph{forward process}, which adds noise to the original data, is defined as
\vspace{-5pt}
\begin{align}
\label{eq:xt}
    \bm{x_t} = \sqrt{\bar{\alpha}}_t \bm{x_0} + \sqrt{1 - \bar{\alpha}_t} \bm{\epsilon}, \; \bm{\epsilon} \sim \mathcal{N}(\bm{0}, \bm{I}),
\end{align}
where $\bm{x_0} \sim q_0(\bm{x})$ is a data sample, $t \in  \{T - 1, \dots, 0\}$, and $\bar{\alpha}_t$ is a noise function that decreases with $t$ (see Figure \ref{fig:cat}). Learning the generative model then corresponds to the \emph{reverse process}, which entails fitting the denoising network $f(\bm{x}_t, t)$ using the (simplified) regression objective \cite{ddpm}:
\vspace{-5pt}
\begin{align}
\label{eq:diff}
    \mathcal{L} = \mathbb{E}_{t, \bm{x_0}, \bm{\epsilon}} ||f(\sqrt{\bar{\alpha}_t} \bm{x_0} + \sqrt{1 - \bar{\alpha}_t} \bm{\epsilon}, t) - \bm{\epsilon} ||^2 \: .
\end{align}
After training, new samples are generated by first sampling $\bm{x_T} \sim \mathcal{N}(\bm{0}, \bm{I})$, and then iteratively applying the denoising network $f$  according to the transition rules from DDPM \cite{ddpm} or DDIM \cite{ddim}. % While we omit the differential equations perspective on diffusion models \cite{song2020score} for brevity, note that our DuoDiff approach can also be applied there.

\textbf{Early-exiting} is a popular paradigm for making inference more efficient by allowing the model's depth to adapt based on the difficulty of the given input \cite{teerapittayanon2016branchynet}. It has been successfully applied across various domains, including computer vision \cite{huang2017multi, jazbec2024towards} and language modeling \cite{elbayad2019depth, schuster2022confident}. For diffusion models, early-exiting has been previously explored in AdaDiff \cite{adadiff}. To enable dynamic inference in AdaDiff, intermediate output heads are attached to the original backbone model (U-ViT \cite{uvit}) before each layer $i=0,\ldots,N - 1$. Furthermore, an uncertainty estimate, $u_{i, t} \in [0,1]
$, is defined at every sampling step $t$ and at each layer $i$. The early (noise) prediction is returned once the uncertainty at a given layer falls below a predefined threshold $\theta \in [0,1]$:
\vspace{-5pt}
\begin{align}
\label{eq:ee}
f(\bm{x_t}, t;\theta) := \begin{cases}
g_0(L_{0, t}) & \text{if } u_{0, t} \leq \theta, \\
\vdots & \vdots \\
g_{N - 1}(L_{{N - 1}, t}) & \text{if } u_{N - 1, t} \leq \theta, \\
g_{N}(L_{{N}, t}) & \text{otherwise.}
\end{cases}
\end{align}
where $L_{i, t}$ denotes the activations before layer $i$, and $g_i$ denotes the $i$-th output head. For more details, refer to Appendix \ref{sec:appendix-models}.

\section{Methods}

% \subsection{Early-Exit Trends} \label{sec:ee-trends}

\paragraph{Early-Exit Trends in Diffusion Models.} 

\begin{figure}
     \centering
     \begin{subfigure}[t]{0.245\textwidth}
         \centering
         \includegraphics[width=\textwidth]{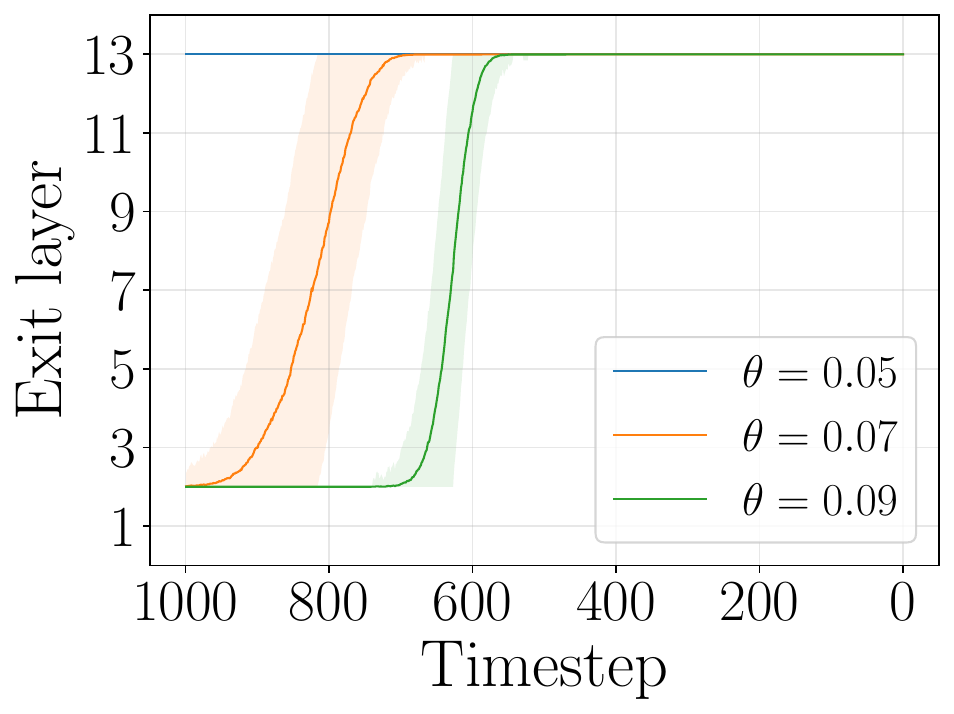}
         \caption{CIFAR-10}
         \label{fig:y equals x}
     \end{subfigure}
     \hfill
     \begin{subfigure}[t]{0.245\textwidth}
         \centering
         \includegraphics[width=\textwidth]{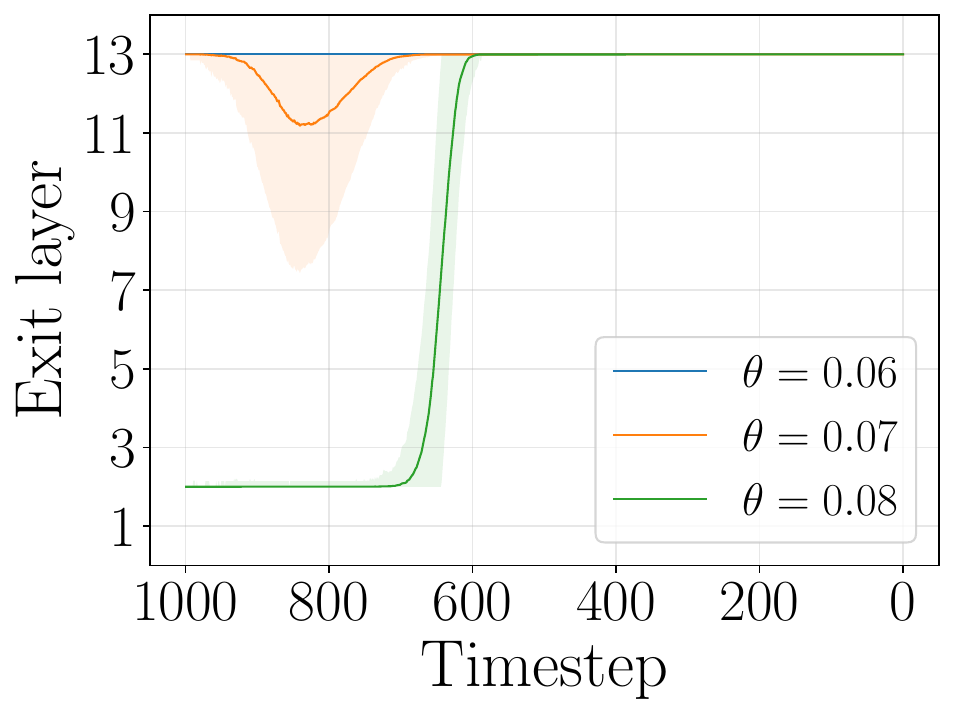}
         \caption{CelebA}
         \label{fig:three sin x}
     \end{subfigure}
     \hfill
     \begin{subfigure}[t]{0.245\textwidth}
         \centering
         \includegraphics[width=\textwidth]{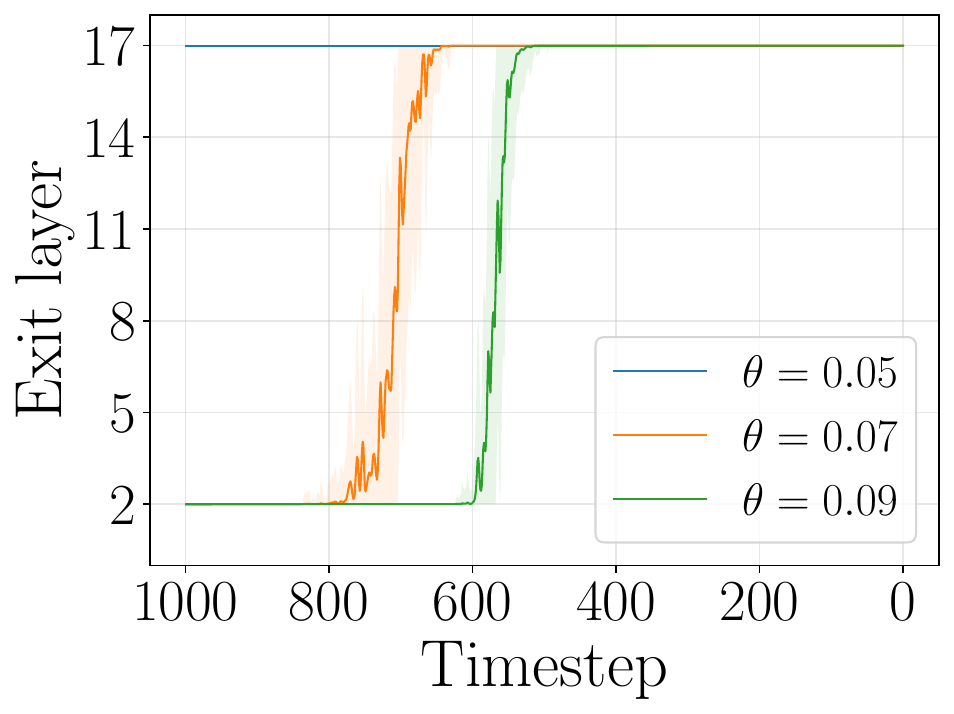}
         \caption{ImageNet ($64 \times 64$)}
         \label{fig:five over x}
     \end{subfigure}
     \hfill
     \begin{subfigure}[t]{0.245\textwidth}
         \centering
         \includegraphics[width=\textwidth]{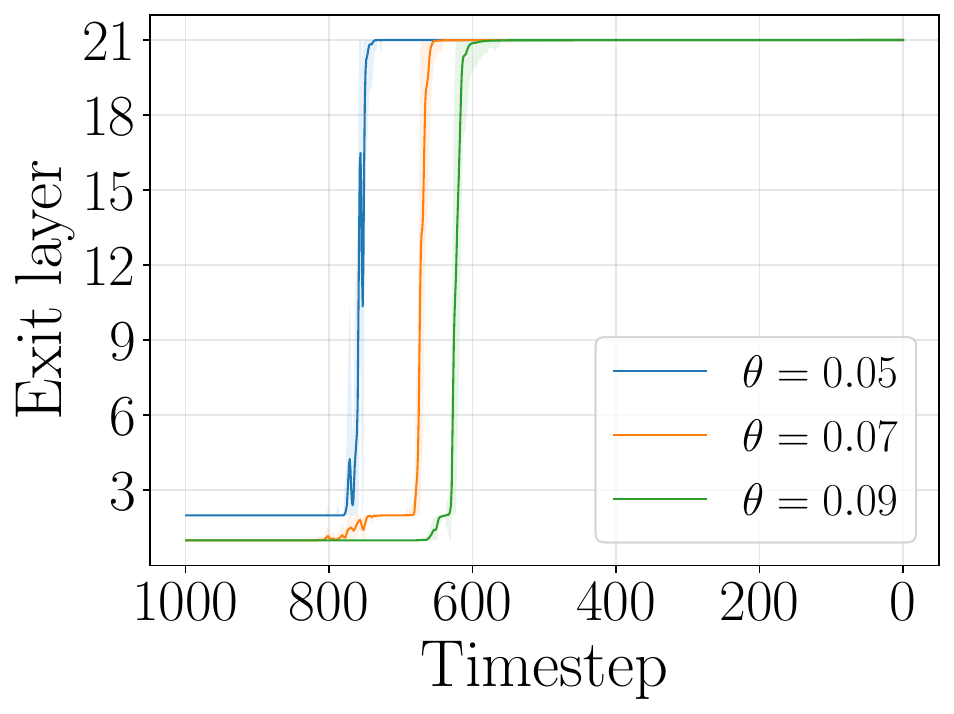}
         \caption{ImageNet ($256 \times 256$)}
         \label{fig:five over x}
     \end{subfigure}
\caption{\textbf{Early-exit trends in AdaDiff \cite{adadiff}.} The plots show the average exit layer across 5,120 images for different datasets and various exiting thresholds $\theta$. We observe that early-exiting in the denoising network occurs only at the start of the generation process (for $t$ close to $T$), followed by a sudden switch to using the full denoising network for the remaining generation steps. The pattern is consistent across different datasets and resembles a step function.} 
        \label{fig:ee-trends}
\end{figure}

%% previous version is commented out below, feel free to revert back if you prefer!!
We begin by leveraging the adaptivity of AdaDiff \cite{adadiff} to study the dynamics of the generative process in diffusion models. Specifically, in Figure \ref{fig:ee-trends}, we visualize the average exit layer across test samples for each sampling step $t$. Interestingly, we observe that early-exit occurs exclusively at the beginning of the reverse process. For example, on ImageNet $64 \times 64$ with threshold $\theta=0.09$, the average exit layer equals 2 until $t\approx600$, after which the full model is utilized. This suggests that, based on AdaDiff's exit trends, the diffusion generative process can be roughly divided into two stages, with the first one being `easier' than the second.

\begin{figure}[t]
     \centering  
     \hspace*{0.1cm}\includegraphics[width=0.67\textwidth]{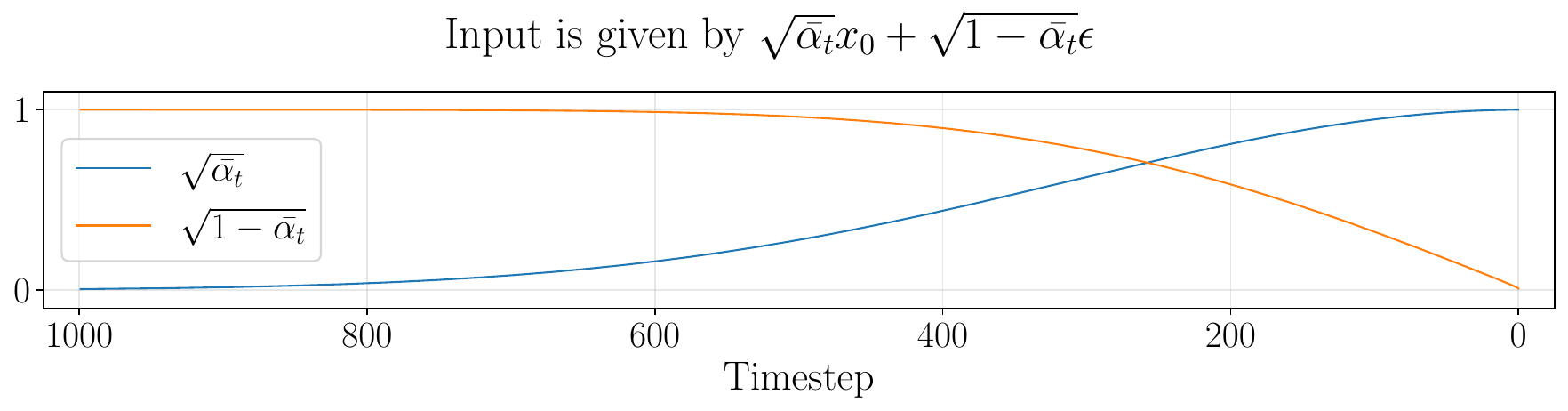}
    \includegraphics[width=0.67\textwidth]{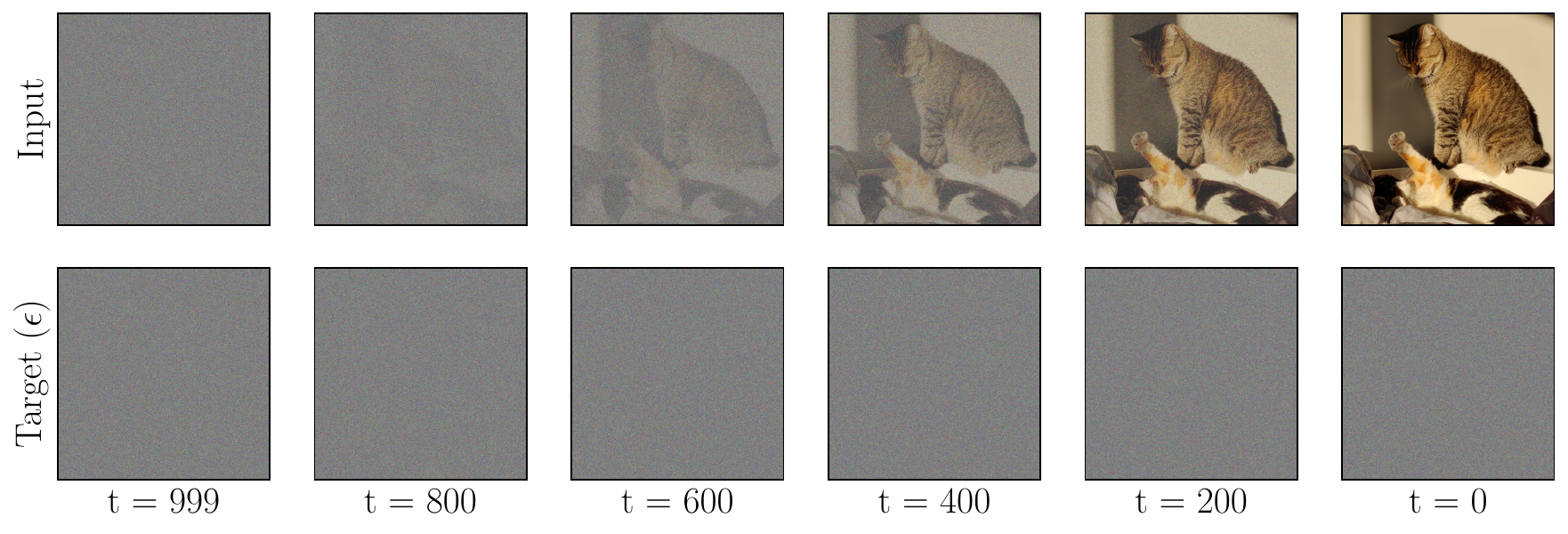}
    \caption{\textbf{Denoising objective.} Given a noisy image and a timestep, the model must predict the added noise. As we can observe, this task is easier for high values of $t$, in which the expected output is very similar to the input.}
     \label{fig:cat}
\end{figure}

Although such behaviour is surprising at first (and was overlooked in the original AdaDiff paper \cite{adadiff}), we demonstrate that it can be explained by taking a closer look at the training of diffusion models. To this end, observe how as $t$ grows, $\bm{x_t}$ becomes increasingly dominated by noise (Eq. \ref{eq:xt}). Consequently, the input and expected output of the denoising network $f$ begin to resemble each other more closely, making the task easier, as the network primarily needs to learn an identity-like behavior \footnote{This is further supported by the training loss being larger for smaller values of $t$, e.g., see Figure 2 in \cite{nichol2021improved}.}. See Figure \ref{fig:cat} for a more visual explanation. This is also reflected at test time, with the denoising network consistently early-exiting for larger $t$, indicating an easier task.

% To better understand the behavior of early-exit during inference, we conducted an in-depth analysis, focusing on identifying the layers at which early-exit occurs throughout the diffusion process. Our observations reveal that models tend to exit very early in the first time steps of the reverse diffusion process, while requiring (nearly) full computation for the final steps. This trend is visually illustrated in Figure \ref{fig:ee-trends}, which shows the early-exit behavior across different datasets.

% This observation is rooted in the theoretical underpinnings of the diffusion process. The model is trained to predict the noise component $\boldsymbol \epsilon$ from a noisy input at each timestep, based on the relation 
% $\mathbf{x}_t = \sqrt{\bar \alpha_t} \mathbf{x}_0 + \sqrt{1 - \bar \alpha _t} \boldsymbol \epsilon$. As $t$ increases, $\sqrt{\bar \alpha_t}$ diminishes and $\sqrt{1 - \bar \alpha _t}$ approaches 1, meaning that at larger t, the model's task simplifies, and its behavior becomes almost identity-like as the input is primarily dominated by noise. This makes the denoising error increase as the diffusion process advances, especially in the latest timesteps \cite{nichol2021improved, adadiff}. 

% Figure \ref{fig:cat} further illustrates this concept. Grounding our observations in this theoretical framework, we hypothesize that during the early steps of the reverse diffusion (when $t$ is large, close to $T$), the model is easily able to predict the noise. As $t$ decreases (closer to 0), the input distribution diverges from that of the noise, and the task becomes increasingly more challenging.

\begin{figure}[t]
     \centering
     \includegraphics[width=0.9\textwidth]{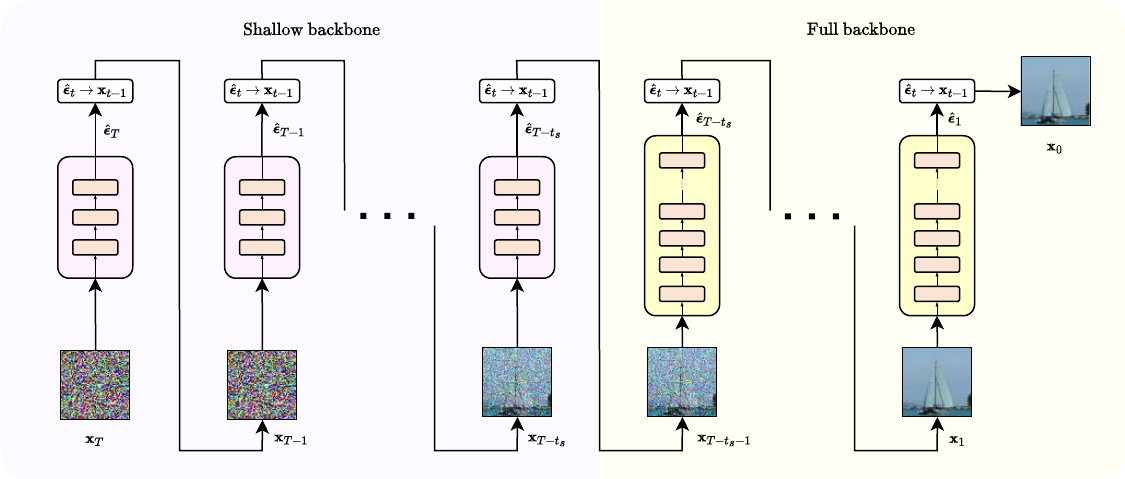}
     % \vspace{8pt}
     \caption{\textbf{DuoDiff framework.} DuoDiff employs a shallow three-layer U-ViT backbone for the first $t_s$ timesteps to reduce computational overhead, before switching to a full backbone for the remaining denoising steps, ensuring both efficiency and image quality. Both backbones are trained on the same dataset using the same diffusion objective.}
     \label{fig:duodiff}
\end{figure}

% \subsection{DuoDiff}
\paragraph{DuoDiff.} Building upon the early-exit trends reported above, we propose DuoDiff, a novel diffusion framework designed to accelerate inference by employing a dual-backbone architecture. During the initial timesteps of the reverse diffusion process, where the input is largely dominated by noise and the task is simpler, DuoDiff utilizes a shallow three-layer backbone, as the early-exit layer for most samples in these timesteps is usually lower than 3 (Figure \ref{fig:ee-trends}). As the diffusion process progresses and the input becomes more structured, DuoDiff switches to the full backbone for the remaining, more complex timesteps. We denote by $t_s$ the number of steps during which the shallow model is active. Both the shallow and the complete backbones are trained from scratch on the same dataset using the same diffusion training objective. In addition, both backbones are trained for all values of $t$ such that one can freely choose $t_s$ after training. Figure \ref{fig:duodiff} provides a detailed illustration of DuoDiff's design.

Unlike AdaDiff, which relies on dynamic early-exit mechanisms based on per-sample uncertainty levels (Eq. \ref{eq:ee}), DuoDiff simplifies this process by using a fixed transition point between the two backbones. While this sacrifices the adaptiveness of early-exiting (i.e., varying compute based on sample's difficulty), we believe this is well justified here as we observe very little variability in exiting patterns between different samples (as indicated by small standard deviation bars in Figure \ref{fig:ee-trends}). Moreover, the static approach eliminates the batching inefficiencies caused by AdaDiff's varying exit points for different samples (see Appendix \ref{sec:batching-issues}), making batch inference more efficient and easier to implement. 

% DuoDiff improves on AdaDiff not only by reducing the computational overhead of the early-exit strategy, but also by making batching simpler. The reason for this is that different samples in a batch might early-exit at different layers, which makes batching hard to implement (see Appendix \ref{sec:batching-issues}). DuoDiff's design effectively eliminates the complexity of managing varying exit points across different samples in a batch, significantly improving the overall efficiency of inference.

\section{Experiments}

In order to illustrate the capabilities of DuoDiff, we compare it to AdaDiff on three widely used datasets: CIFAR-10 $32 \times 32$ \cite{cifar10} and CelebA $64 \times 64$ \cite{celeba} for unconditional generation and ImageNet \cite{imagenet} for class-conditional generation. For ImageNet, we evaluate the models on two resolutions: $64 \times 64$ and $256 \times 256$, enabling us to assess DuoDiff's scalability across varying image sizes. For ImageNet $256 \times 256$, we train our diffusion models in latent space. We utilize the U-ViT \cite{uvit} architecture as the base model. In all experiments, DuoDiff employs a shallow three-layer backbone, while the full model varies in size depending on the dataset (see Tables \ref{tab:results} and \ref{tab:hp-model}).

We evaluate the quality of the generated images using the FID score \cite{fid} and measure the performance by recording the inference time per sample. Additionally, we test DuoDiff with both DDPM and DDIM samplers and provide evidence that DuoDiff works seamlessly with latent space diffusion. All metrics are computed over 5,120 images, processed in batches of 128. For AdaDiff, computing the inference time using batch sampling is challenging. For more details on batching, see Appendix \ref{sec:batching-issues}. 

Appendix \ref{sec:model-specifications} presents the hyperparameters and further implementation details. We also make publicly available our code on GitHub\footnote{\url{https://github.com/razvanmatisan/duodiff}} which contains both the DuoDiff and AdaDiff implementations together with experiments, configuration files, and demo notebooks.

\paragraph{Performance and Image Quality on AdaDiff.}

\begin{figure}[]
     \centering
     \includegraphics[width=0.43\textwidth]{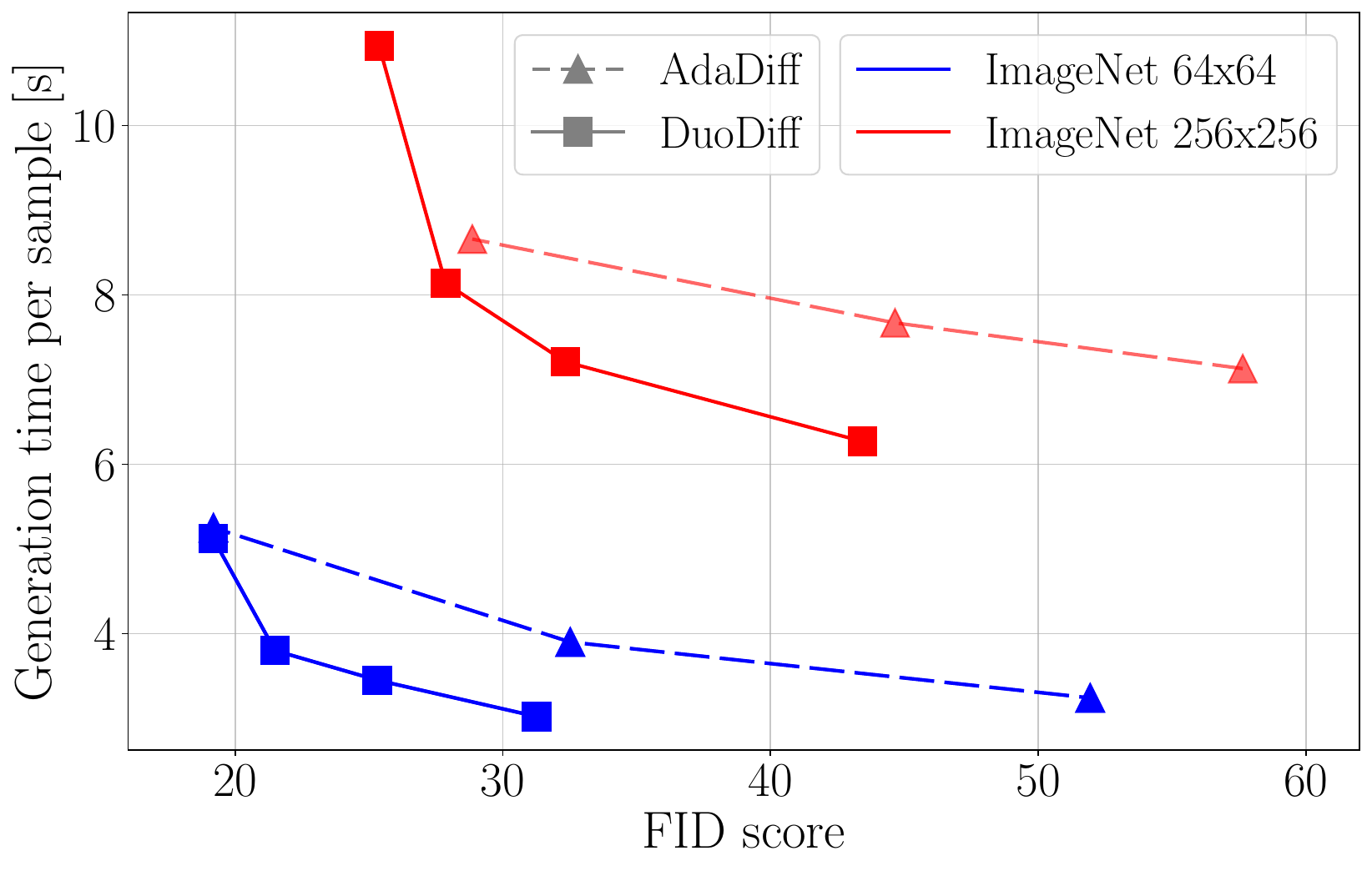}
     \caption{\textbf{Comparison of AdaDiff and DuoDiff.} Comparison of AdaDiff and DuoDiff. The plot shows FID score and generation time per sample (lower is better for both) across two datasets (ImageNet $64\times64$ and $256\times256$). Each point represents a different parameter configuration, including the base model, which can be seen as a special case of DuoDiff ($t_s = 0$). We can see how DuoDiff consistently outperforms AdaDiff in both performance and inference time. } 
     \label{fig:imagenet}
\end{figure}

In this study, we compare the performance of AdaDiff and DuoDiff, demonstrating that DuoDiff surpasses AdaDiff in both image quality and sampling efficiency. Figure \ref{fig:imagenet} illustrates the FID scores and inference time across ImageNet $64 \times 64$ and ImageNet $256 \times 256$. For a tabular view of all the results, please refer to Appendix \ref{sec:quantitative-results-appendix}.

DuoDiff demonstrates superior performance over both the baseline and AdaDiff in terms of inference time. This outcome is expected, as DuoDiff leverages a shallow U-ViT for the first timesteps, while AdaDiff incurs additional overhead from its uncertainty-based early-exit mechanism. 

However, the decrease in sampling time for both methods is accompanied by a decline in FID scores. For AdaDiff, this decline is more pronounced, with a clear trade-off between faster inference and lower image quality as $\theta$ increases. In contrast, while DuoDiff also experiences a reduction in FID scores as the $t_s$ value increases, this decline is significantly less severe compared to AdaDiff, with the image quality remaining more stable and closer to the baseline. For example, on ImageNet $256 \times 256$ and with a computational budget of $7s$ per sample, AdaDiff achieves a FID score of $57$, whereas DuoDiff achieves a FID score of $32$ -- an improvement of roughly $40\%$. Refer to Table \ref{tab:results} for more details and quantitative results.

Moreover, Table \ref{table:latent-diffusion_and_ddim} demonstrates that DuoDiff can be used alongside other techniques such as DDIM \cite{ddim} and latent diffusion \cite{rombach2022high}. 

\begin{table}[]
\caption{\textbf{Compatibility with DDIM.} Image quality (FID score) and inference speed using DuoDiff with DDIM sampling in a latent space. We observe how DuoDiff successfully increases the sampling speed without a significant impact in image quality.}
\label{table:latent-diffusion_and_ddim}
\small
\centering
\begin{tabular}{lllll}
\textbf{Dataset} & \textbf{Base model} & \textbf{Inference method} & \textbf{FID score $\downarrow$} & \textbf{Inference Time [s] $\downarrow$} \\ \hline
\multirow{4}{*}{\makecell{ImageNet \\ ($256 \times 256$)}} 
    & \multirow{4}{*}{U-ViT-L/2} 
    & DDIM ($\eta = 0, n\_steps = 50$) & 27.82 & 0.55 \\ 
    &  & \, + DuoDiff ($t_s = 150$) & 29.17 & 0.47 \\
    &  & \, + DuoDiff ($t_s = 200$) & 30.06 & 0.46 \\
    &  & \, + DuoDiff ($t_s = 300$) & 34.36 & 0.41 \\ \hline
\end{tabular}
\normalsize
\end{table}

\paragraph{Hyperparameters Effect on Performance.}

\begin{figure}[]
\centering
    % First subfigure
    \begin{subfigure}[b]{0.49\textwidth}
        \centering
        \includegraphics[width=\textwidth]{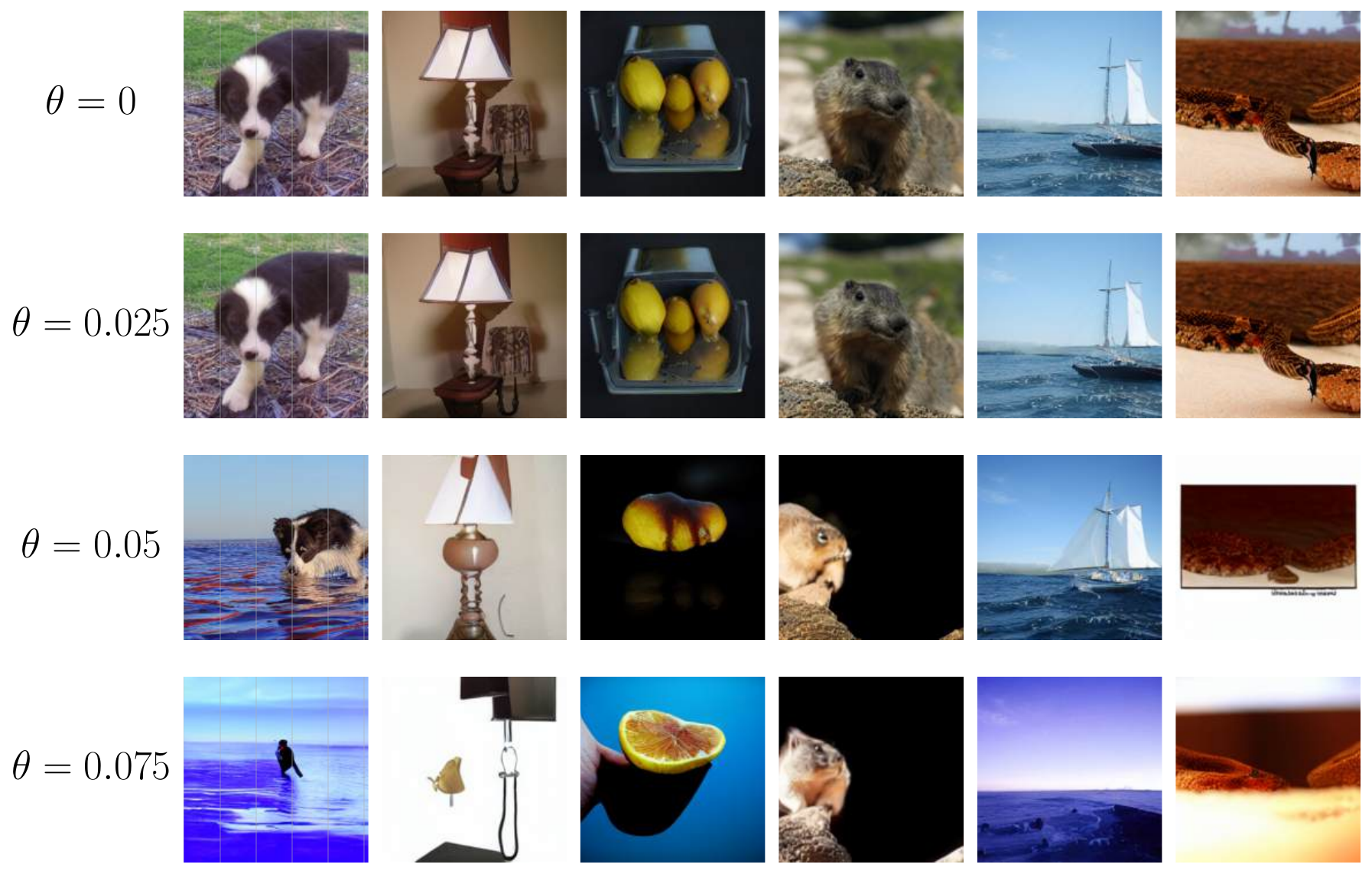}
        \caption{AdaDiff threshold ($\theta$)}
        \label{fig:threshold_1}
    \end{subfigure}
    \hfill
    % Second subfigure
    \begin{subfigure}[b]{0.49\textwidth}
        \centering
        \includegraphics[width=\textwidth]{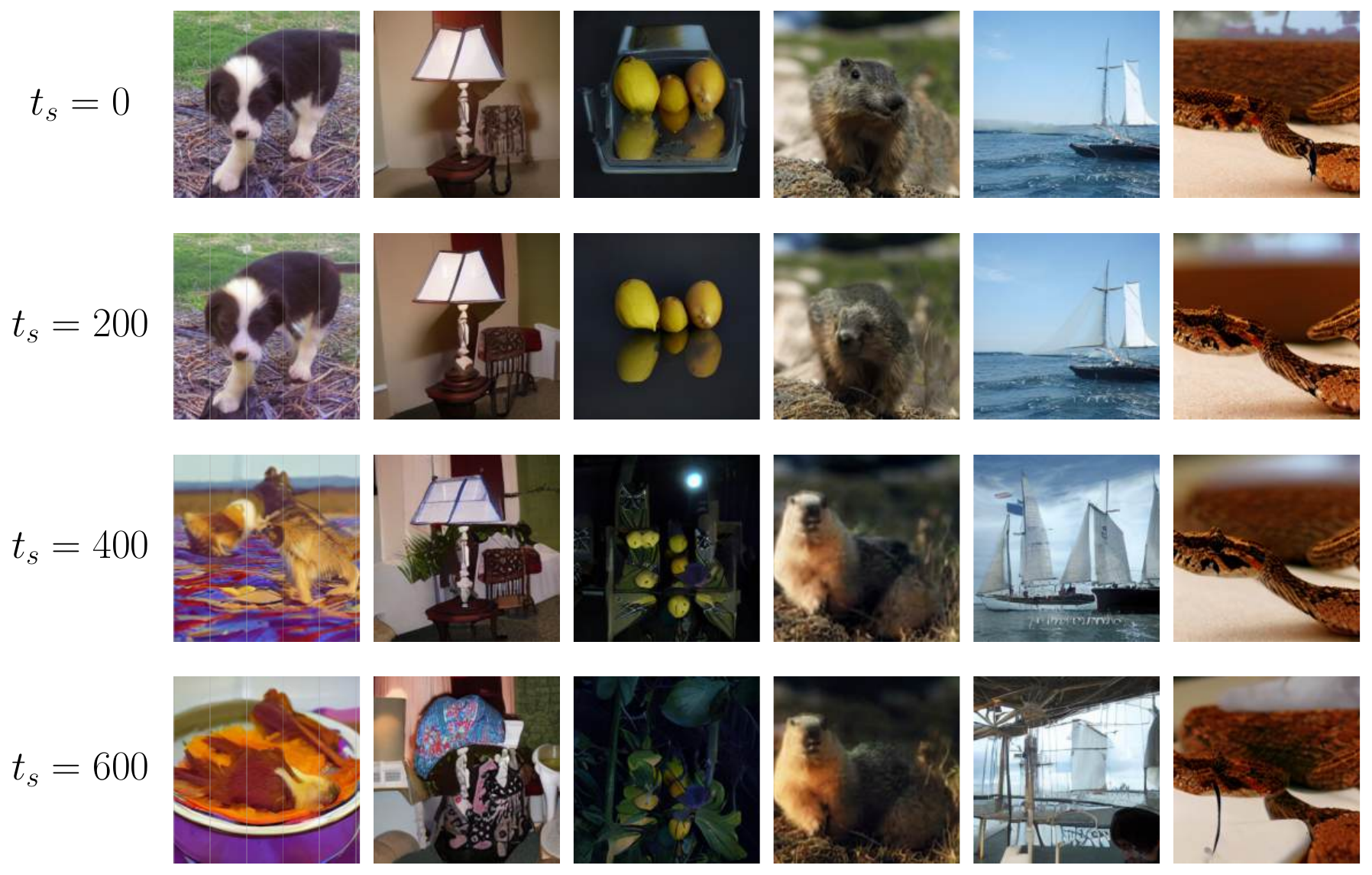}
        \caption{DuoDiff switch time ($t_s$)}
        \label{fig:threshold_2}
    \end{subfigure}
    \caption{\textbf{Qualitative hyperparameter analysis.} Comparison of image generation results for AdaDiff (left) and DuoDiff (right) on the ImageNet dataset ($256 \times 256$) using different values for their respective hyperparameters ($\theta$ in AdaDiff and $t_s$ in DuoDiff). We observe how higher values of $\theta$ and $t_s$ diminish the quality of the generated images.}
    \label{fig:qualitative_results}
\end{figure}

A general trend can be observed for both AdaDiff and DuoDiff: as the threshold hyperparameters ($\theta$ and $t_s$, respectively) increase, image quality progressively degrades, while inference time decreases. This relationship is illustrated qualitatively in Figure \ref{fig:qualitative_results} and quantified in Table \ref{tab:results}. We leave for future work the incorporation of more principled mechanisms for threshold selection.

% To determine the optimal values of these hyperparameters, we suggest using a small calibration dataset and finding the smallest values for which the performance drop remains negligible. We leave the incorporation of more principled selection mechanisms \cite{jazbec2024fast} for future work. 

\section{Conclusion \& Future Work}
In this paper, we have introduced DuoDiff, a dual-backbone alternative to adaptive diffusion models motivated by the consistency of the early-exit trends. We show that DuoDiff substantially decreases per-sample inference time while maintaining image quality. DuoDiff is also compatible with other diffusion techniques, including latent space diffusion and DDIM sampling, providing an efficient solution to address the slow inference speed of diffusion models. 

Future research will focus on exploring different DuoDiff configurations, such as increasing the number of layers in the shallow transformer in order to increase $t_s$. Additionally, a promising direction involves investigating early-exit trends across different diffusion parametrizations, such as predicting the original image rather than the added noise. 

% \bibliography{references}
% \bibliographystyle{bibstyle}

\newpage

\printbibliography

@article{ddim,
  title={Denoising diffusion implicit models},
  author={Song, Jiaming and Meng, Chenlin and Ermon, Stefano},
  journal={arXiv preprint arXiv:2010.02502},
  year={2020}
}

@article{ddpm,
  title={Denoising diffusion probabilistic models},
  author={Ho, Jonathan and Jain, Ajay and Abbeel, Pieter},
  journal={Advances in neural information processing systems},
  volume={33},
  pages={6840--6851},
  year={2020}
}

@inproceedings{uvit,
  title={All are worth words: A vit backbone for diffusion models},
  author={Bao, Fan and Nie, Shen and Xue, Kaiwen and Cao, Yue and Li, Chongxuan and Su, Hang and Zhu, Jun},
  booktitle={Proceedings of the IEEE/CVF conference on computer vision and pattern recognition},
  pages={22669--22679},
  year={2023}
}

@article{distil1,
  title={Knowledge distillation in iterative generative models for improved sampling speed},
  author={Luhman, Eric and Luhman, Troy},
  journal={arXiv preprint arXiv:2101.02388},
  year={2021}
}

@article{distil2,
  title={Progressive distillation for fast sampling of diffusion models},
  author={Salimans, Tim and Ho, Jonathan},
  journal={arXiv preprint arXiv:2202.00512},
  year={2022}
}

@Techreport{cifar10,
 author = {Krizhevsky, Alex and Hinton, Geoffrey},
 address = {Toronto, Ontario},
 institution = {University of Toronto},
 number = {0},
 publisher = {Technical report, University of Toronto},
 title = {Learning multiple layers of features from tiny images},
 year = {2009},
 title_with_no_special_chars = {Learning multiple layers of features from tiny images},
 url = {https://www.cs.toronto.edu/~kriz/learning-features-2009-TR.pdf}
}

@inproceedings{celeba,
  author={Liu, Ziwei and Luo, Ping and Wang, Xiaogang and Tang, Xiaoou},
  booktitle={ICCV}, 
  title={Deep Learning Face Attributes in the Wild}, 
  year={2015},
  volume={},
  number={}
}

@INPROCEEDINGS{imagenet,
  author={Deng, Jia and Dong, Wei and Socher, Richard and Li, Li-Jia and Kai Li and Li Fei-Fei},
  booktitle={2009 IEEE Conference on Computer Vision and Pattern Recognition}, 
  title={ImageNet: A large-scale hierarchical image database}, 
  year={2009},
  volume={},
  number={},
  pages={248-255},
  doi={10.1109/CVPR.2009.5206848}}

@article{fid,
  title={Gans trained by a two time-scale update rule converge to a local nash equilibrium},
  author={Heusel, Martin and Ramsauer, Hubert and Unterthiner, Thomas and Nessler, Bernhard and Hochreiter, Sepp},
  journal={NeurIPS},
  year={2017}
}

@misc{adadiff,
      title={AdaDiff: Accelerating Diffusion Models through Step-Wise Adaptive Computation}, 
      author={Shengkun Tang and Yaqing Wang and Caiwen Ding and Yi Liang and Yao Li and Dongkuan Xu},
      year={2024},
      eprint={2309.17074},
      archivePrefix={arXiv}, 
}

@inproceedings{sohl2015deep,
  title={Deep unsupervised learning using nonequilibrium thermodynamics},
  author={Sohl-Dickstein, Jascha and Weiss, Eric and Maheswaranathan, Niru and Ganguli, Surya},
  booktitle={International conference on machine learning},
  pages={2256--2265},
  year={2015},
  organization={PMLR}
}

@article{dhariwal2021diffusion,
  title={Diffusion models beat gans on image synthesis},
  author={Dhariwal, Prafulla and Nichol, Alexander},
  journal={Advances in neural information processing systems},
  volume={34},
  pages={8780--8794},
  year={2021}
}

@inproceedings{rombach2022high,
  title={High-resolution image synthesis with latent diffusion models},
  author={Rombach, Robin and Blattmann, Andreas and Lorenz, Dominik and Esser, Patrick and Ommer, Bj{\"o}rn},
  booktitle={Proceedings of the IEEE/CVF conference on computer vision and pattern recognition},
  pages={10684--10695},
  year={2022}
}

@inproceedings{teerapittayanon2016branchynet,
  title={Branchynet: Fast inference via early exiting from deep neural networks},
  author={Teerapittayanon, Surat and McDanel, Bradley and Kung, Hsiang-Tsung},
  booktitle={2016 23rd international conference on pattern recognition (ICPR)},
  pages={2464--2469},
  year={2016},
  organization={IEEE}
}

@inproceedings{moon2023early,
  title={Early exiting for accelerated inference in diffusion models},
  author={Moon, Taehong and Choi, Moonseok and Yun, EungGu and Yoon, Jongmin and Lee, Gayoung and Lee, Juho},
  booktitle={ICML 2023 Workshop on Structured Probabilistic Inference $\{$$\backslash$\&$\}$ Generative Modeling},
  year={2023}
}

@article{ho2022video,
  title={Video diffusion models},
  author={Ho, Jonathan and Salimans, Tim and Gritsenko, Alexey and Chan, William and Norouzi, Mohammad and Fleet, David J},
  journal={Advances in Neural Information Processing Systems},
  volume={35},
  pages={8633--8646},
  year={2022}
}

@article{ho2022imagen,
  title={Imagen video: High definition video generation with diffusion models},
  author={Ho, Jonathan and Chan, William and Saharia, Chitwan and Whang, Jay and Gao, Ruiqi and Gritsenko, Alexey and Kingma, Diederik P and Poole, Ben and Norouzi, Mohammad and Fleet, David J and others},
  journal={arXiv preprint arXiv:2210.02303},
  year={2022}
}

@article{kong2020diffwave,
  title={Diffwave: A versatile diffusion model for audio synthesis},
  author={Kong, Zhifeng and Ping, Wei and Huang, Jiaji and Zhao, Kexin and Catanzaro, Bryan},
  journal={arXiv preprint arXiv:2009.09761},
  year={2020}
}

@inproceedings{hoogeboom2022equivariant,
  title={Equivariant diffusion for molecule generation in 3d},
  author={Hoogeboom, Emiel and Satorras, V{\i}ctor Garcia and Vignac, Cl{\'e}ment and Welling, Max},
  booktitle={International conference on machine learning},
  pages={8867--8887},
  year={2022},
  organization={PMLR}
}

@book{tomczak2022deep,
title = "Deep Generative Modeling",
author = "Tomczak, {Jakub M.}",
year = "2022",
month = feb,
doi = "10.1007/978-3-030-93158-2",
language = "English",
isbn = "978-3-030-93157-5",
publisher = "Springer",
address = "Germany",
}

@article{ulhaq2022efficient,
  title={Efficient diffusion models for vision: A survey},
  author={Ulhaq, Anwaar and Akhtar, Naveed and Pogrebna, Ganna},
  journal={arXiv preprint arXiv:2210.09292},
  year={2022}
}

@article{nichol2021improved,
  author       = {Alex Nichol and
                  Prafulla Dhariwal},
  title        = {Improved Denoising Diffusion Probabilistic Models},
  journal      = {CoRR},
  year         = {2021},
}

@article{huang2017multi,
  title={Multi-scale dense networks for resource efficient image classification},
  author={Huang, Gao and Chen, Danlu and Li, Tianhong and Wu, Felix and Van Der Maaten, Laurens and Weinberger, Kilian Q},
  journal={arXiv preprint arXiv:1703.09844},
  year={2017}
}

@article{jazbec2024towards,
  title={Towards anytime classification in early-exit architectures by enforcing conditional monotonicity},
  author={Jazbec, Metod and Allingham, James and Zhang, Dan and Nalisnick, Eric},
  journal={Advances in Neural Information Processing Systems},
  volume={36},
  year={2024}
}

@article{elbayad2019depth,
  title={Depth-adaptive transformer},
  author={Elbayad, Maha and Gu, Jiatao and Grave, Edouard and Auli, Michael},
  journal={arXiv preprint arXiv:1910.10073},
  year={2019}
}

@article{schuster2022confident,
  title={Confident adaptive language modeling},
  author={Schuster, Tal and Fisch, Adam and Gupta, Jai and Dehghani, Mostafa and Bahri, Dara and Tran, Vinh and Tay, Yi and Metzler, Donald},
  journal={Advances in Neural Information Processing Systems},
  volume={35},
  pages={17456--17472},
  year={2022}
}

\newpage
\appendix
\onecolumn
 \section{AdaDiff \label{sec:appendix-models}}
\subsection{Architecture}

\begin{figure}[ht]
    \centering
    % First subfigure
    \begin{subfigure}[b]{0.70\textwidth}
        \centering
        \includegraphics[width=\textwidth]{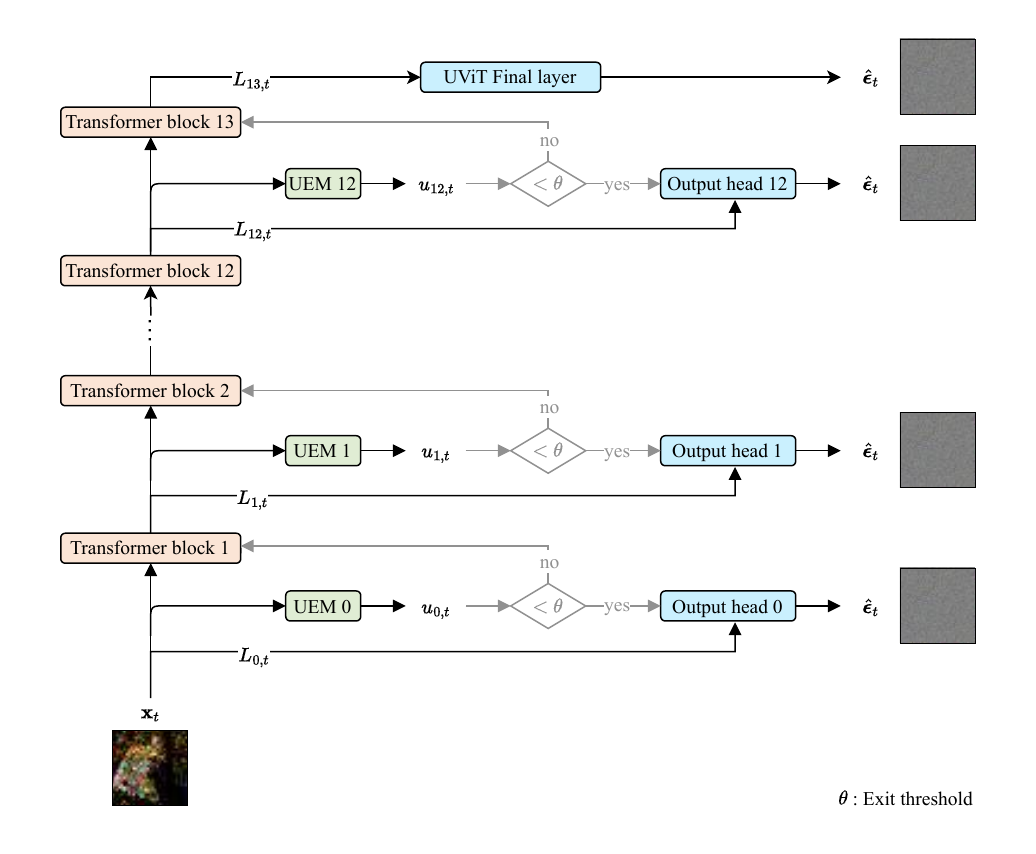} 
        \caption{Architecture}
        \label{fig:architecture}
    \end{subfigure}
    \quad
    % Second subfigure
    \begin{subfigure}[b]{0.26\textwidth}
        \centering
        \raisebox{1.5cm}{
        \includegraphics[width=\textwidth]{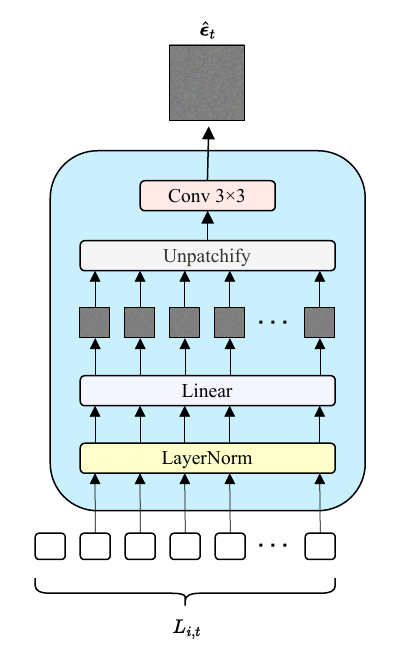}}
        \caption{Output head}
        \label{fig:outputhead}
    \end{subfigure}
    \caption{\textbf{AdaDiff architecture.} AdaDiff architecture integrated in a U-ViT transformer with 13 layers. An Uncertainty Estimation Module is included before each transformer block to check whether early-exiting can be applied. In the affirmative case, an output head computes the predicted noise from the output of the previous transformer block. U-ViT skip connections are omitted for simplicity.}
    \label{fig:maindiagram}
\end{figure}

AdaDiff implements a dynamic early-exit strategy, where Uncertainty Estimation Modules (UEMs) are used to determine whether computation can be halted at each layer of the model. This process is illustrated in Figure \ref{fig:maindiagram}, which displays the AdaDiff architecture built on top of a 13-layer U-ViT transformer, and the architectural design of the output heads.

% \begin{figure}[H]
%   \centering
%   \includegraphics[width=0.6\textwidth]{figures/combined_diagram.png}
%   \caption{AdaDiff architecture integrated in a U-ViT-Small transformer. U-ViT skip connections are ommited for simplicity.}
%   \label{fig:diagram}
% \end{figure}

\subsection{Timestep-Aware Uncertainty Estimation Module (UEM)}
For the implementation of the uncertainty estimation networks, they propose a timestep-aware UEM in the form of a fully-connected layer:

\begin{equation}
u_{i, t}=f\left(\mathbf{w}_ {t}^{T}\left[L_ {i, t}\right.\right., timesteps \left.]+\mathbf b_ {t}\right)
\end{equation} 

where $\mathbf w_t$, $\mathbf b_t$, $f$, and $timesteps$ are the weight matrix, weight bias, activation function, and timestep embeddings, respectively. The pseudo-uncertainty ground truth is constructed as follows:

\begin{equation}
    \hat{u}_ {i, t}=F\left(\left|\mathbf{g}_ {i}\left(L_{i, t}\right)-\boldsymbol \epsilon\right|\right)
\end{equation}

where $\mathbf{g}_i$ is the output head, $\bm{\epsilon}$ is the ground truth noise value and $F$ is a function to keep the output smaller than one (the authors use $F = \tanh$). The implementation of the output layer, shown in Figure \ref{fig:outputhead}, is inspired on the final layer of the U-ViT architecture.This brings forth the loss function of this module, designed as the MSE loss of the estimated and pseudo-uncertainty ground truth:

\begin{equation}
    \mathcal{L}_ {u}^{t}=\sum_{i=0}^{N-1}\left\|u_{i, t}-\hat{u}_{i, t}\right\|^{2}.
\end{equation}

\begin{figure}[ht]
    \centering
        \centering
        \includegraphics[width=0.35\textwidth]{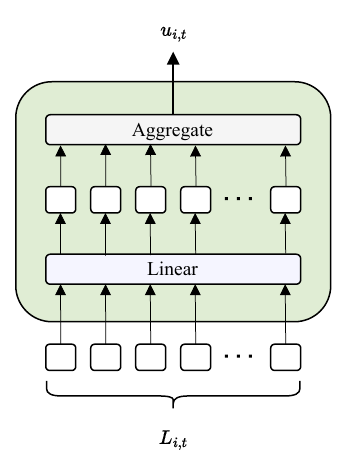} 
        \caption{\textbf{Uncertainty Estimation Module.} We design the UEM as an multilayer perceptron, as specified by the AdaDiff's authors.}
        \label{fig:uem}
\end{figure}

During inference, early-exiting is then achieved by comparing the estimated uncertainty of the output prediction from each layer with a predefined threshold. Figure \ref{fig:uem} provides a visual representation of the UEM.

\subsection{Uncertainty-Aware Layer-wise Loss}

The authors also propose an uncertainty-aware layer-wise loss. They draw inspiration from previous work, with one important modification, a weighting term to give more importance to the output layers where the uncertainty is lower (i.e., early-exiting will happen).
\begin{equation}
    \mathcal{L}_ {U A L}^{t}=\sum_{i=0}^{N-1}\left(1-u_{i, t}\right) \times\left\|\mathbf{g}_ {i}\left(L_{i, t}\right)-\boldsymbol \epsilon_t\right\|^{2}.
\end{equation}

\subsection{Training Strategy}

AdaDiff utilizes a joint training strategy to balance the effect between uncertainty estimation loss and uncertainty-aware layer-wise loss, added to the orignal diffusion loss:

\begin{equation}
    \mathcal{L}_ {\text {all }}=\mathcal{L}_ {\text {simple }}^{t}(\boldsymbol{\theta})+\lambda \mathcal{L}_ {u}^{t}+\beta \mathcal{L}_{U A L}^{t}
\end{equation}

In their experiments, the authors chose $\lambda = 1$ and $\beta = 1$, which we keep the same throughout our study.

% When we train AdaDiff, we initialize the backbone with pre-trained weights from [...], and keeping it frozen while training the UEMs and the projection heads.

% We use the AdamW \cite{adamw} optimizer with a learning rate of $2 \cdot 10^{-4}$ and a cosine scheduler with a warmup (see our GitHub repo for more details). We train different models by employing the following strategies:
% (1) jointly training the U-ViT, the UEMs and the projection heads from scratch,
% (2) initializing the backbone with pretrained weights, and fine-tuning it while training the UEMs and the projection heads and
% (3) initializing the backbone with pretrained weights, and keeping it frozen while training the UEMs and the projection heads. We pretrained our own weights, which produced better results than the ones available in the U-ViT repository \footnote{https://github.com/baofff/U-ViT}.
% For initializing the backbone, we experiment both with available [pre-trained weights](https://github.com/baofff/U-ViT) as well as with training the U-ViT model ourselves. Following a series of preliminary tests, we established that our weights produce better results overall. Consequently, for all subsequent experiments - except the ones where we jointly train everything from scratch - we used these weights as the starting point.

\subsection{Batching Issues}
\label{sec:batching-issues}

Implementing early-exiting is problematic when the batch size is larger than one, as some samples are ``ready'' to exit early while others are not. A possible implementation would be to use a \textit{shrinking batch size}: start with a fixed batch size (e.g., 128) and as it goes through the transformer, take out the samples that are ready.

To simplify, we \textit{simulated} early-exiting: we make all samples go through the entire transformer, and keep the intermediate activations. Then, we compute where each sample would have exited, and replace the output with the corresponding intermediate activations. Thus, the output is the same as if we had used the \textit{shrinking batch size} implementation.

We also record the exit-layer per sample in order to approximate the inference time. Since all layers and uncertainty modules are identical, we linearly interpolate the total running time using the average exit layer. Note that the \textit{shrinking batch size} implementation would likely result in longer inference times, as it would need to use batch sizes that are not powers of two. Thus, we are underestimating the inference time for AdaDiff.

\section{Quantitative results} \label{sec:quantitative-results-appendix}

Table \ref{tab:results} shows the FID score and inference time obtained for all experiments.

\begin{table}[H]
\caption{\textbf{Quantitative image generation results.} Image generation quality and speed results for the different datasets. $^*$Linearly interpolating with mean exit layer. In practice, it is hard to apply early-exiting with a large batch size. More details regarding the computation of inference time can be found in Appendix \ref{sec:batching-issues} \\}.
\label{tab:results}
\begin{tabular}{lllll}
\textbf{Dataset} & \textbf{Base model} & \textbf{Inference method} & \textbf{FID score $\downarrow$} & \textbf{Inference time [s] $\downarrow$} \\ \hline
\multirow{7}{*}{\makecell{CIFAR10 \\ ($32 \times 32$)}}                  
& \multirow{7}{*}{U-ViT-S/2} & DDPM    & 17.89 & 1.88 \\
                                         &                            & \, + AdaDiff ($\theta = 0.05$) & 17.89 & 1.93 \\
                                         &                            & \, + AdaDiff ($\theta = 0.07$) & 17.55 & $1.63^*$ \\ % 11.0350
                                         &                            & \, + AdaDiff ($\theta = 0.09$) & 24.60 & $1.32^*$ \\ % 8.9205
                                         &                            & \, + DuoDiff ($t_s = 300$)    & 17.81 & 1.45  \\
                                         &                            & \, + DuoDiff ($t_s = 400$)    & 17.95 & 1.30  \\
                                         &                            & \, + DuoDiff ($t_s = 500$)    & 18.67 & 1.16  \\ \hline
\multirow{7}{*}{\makecell{CelebA \\ ($64 \times 64$)}} 
& \multirow{7}{*}{U-ViT-S/4} & DDPM    & 9.98  & 1.88 \\
                                         &                            & \, + AdaDiff ($\theta = 0.06$) & 9.75 & $1.96^*$ \\ % 1.95, 12.7866
                                         &                            & \, + AdaDiff ($\theta = 0.07$) & 9.99 & $1.92^*$ \\ % 1.95, 12.7866
                                         &                            & \, + AdaDiff ($\theta = 0.08$) & 31.41 & $1.36^*$ \\ % 1.95, 9.0926
                                         % &                            & \, + AdaDiff ($\theta = 0.085$) & 45.67 & $^*$ \\ % 1.95, 9.0926

                                         &                            & \, + DuoDiff ($t_s = 300$)    & 10.08 & 1.45  \\ 
                                         &                            & \, + DuoDiff ($t_s = 400$)   & 10.61 & 1.30  \\ 
                                         &                            & \, + DuoDiff ($t_s = 500$)   & 12.18 & 1.16  \\ \hline
\multirow{7}{*}{\makecell{ImageNet \\ ($64 \times 64$)}} 
& \multirow{7}{*}{U-ViT-M/4} & DDPM    & 19.19 & 5.12 \\
                                         &                            & \, + AdaDiff ($\theta = 0.05$) & 19.19 & 5.25 \\ % 5.25
                                         &                            & \, + AdaDiff ($\theta = 0.07$) & 32.52 & $3.90^*$ \\ % 5.25
                                         &                            & \, + AdaDiff ($\theta = 0.09$) & 51.94 & $3.24^*$ \\ % 5.25
                                         &                            & \, + DuoDiff ($t_s = 300$)  & 21.49 & 3.86 \\ 
                                         &                            & \, + DuoDiff ($t_s = 400$)  & 25.31 & 3.45 \\ 
                                         &                            & \, + DuoDiff ($t_s = 500$)  & 31.26 & 3.02 \\ \hline
\multirow{7}{*}{\makecell{ImageNet \\ ($256 \times 256$)}} 
& \multirow{7}{*}{U-ViT-L/2} & DDPM    & 25.38 & 10.94 \\
%                                          &                            & AdaDiff (No EE) & - & 11.09 \\
                                        &                            & \, + AdaDiff ($\theta = 0.05$) & 28.86 & $8.66^*$ \\
                                         &                            & \, + AdaDiff ($\theta = 0.07$) & 44.65 & $7.67^*$ \\
                                         &                            & \, + AdaDiff ($\theta = 0.09$) & 57.64 & $7.13^*$ \\
                                         &                            & \, + DuoDiff ($t_s = 300$)   & 27.86 & 8.14 \\ 
                                         &                            & \, + DuoDiff ($t_s = 400$)   & 32.34 & 7.21 \\
                                         &                            & \, + DuoDiff ($t_s = 500$)   & 43.43 & 6.27 \\ \hline
\multirow{4}{*}{\makecell{ImageNet \\ ($256 \times 256$)}}  & \multirow{4}{*}{U-ViT-L/2} & DDIM ($\eta = 0, n\_steps = 50$) & 27.82 & 0.55 \\ 
 &  & \, + DuoDiff ($t_s = 150$) & 29.17 & 0.47 \\
 &  & \, + DuoDiff ($t_s = 200$) & 30.06  & 0.46 \\
 &  & \, + DuoDiff ($t_s = 300$) & 34.36 & 0.41 \\ \hline
\end{tabular}
\end{table}

% \section{Qualitative results}
% \label{sec:more_qualitative_results}

\section{Model specifications}
\label{sec:model-specifications}

Inspired by the authors of U-ViT, we use the 13-layer configuration for CIFAR-10 (U-ViT-S/2) and CelebA (U-ViT-S/4), as well as 17-layer (U-ViT-M/4) and 21-layer (U-ViT-L/2) configurations for the $64 \times 64$ and $256 \times 256$ ImageNet datasets, respectively. We train everything on a single 40 GB Nvidia A100 GPU except for the full-models for ImageNet, for which we used the weights made public by the authors \cite{uvit}.

The training loss and strategy for AdaDiff are presented in Appendix \ref{sec:appendix-models}. In our experiments, we keep the backbone frozen and train just the output heads and UEMs, as it yielded better performance.

For DuoDiff, we train two U-ViT backbones: a shallow U-ViT with just three layers, and a large one (its size depends on the dataset they were trained, as described previously in this section). The two backbones are trained independently and for all values of $t$. This is important so we can freely decide $t_s$ after training, and to ensure a smooth transition between models. For the ImageNet $256 \times 256$ dataset, we perform diffusion in latent space rather than directly in pixel space due to the large size of the images, which significantly reduces computational overhead. We use a pre-trained autoencoder to map images into latent space, which remains frozen during training. Additionally, we experiment with both DDPM and DDIM samplers, evaluating DuoDiff's performance in terms of inference speed and image quality with each approach.

In Tables \ref{tab:hp-model} and \ref{tab:hp-training}, we present a comprehensive list of the hyperparameters that we used in our experiments.

\begin{table}[h]
\caption{\textbf{U-ViT configurations.} Hyperparameters of the U-ViT backbones. We used a different backbone depending on the dataset and image resolutions used, similar to the official implementation of U-ViT \cite{uvit}. $^*$For DuoDiff, the shallow backbone will have the same model specifications except for the number of layers, which is 3. \\}

\label{tab:hp-model}
\centering

\begin{tabular}{l|l|l|l|l}
% \hline
 & \textbf{CIFAR-10}                           & \textbf{CelebA}     & \textbf{ImageNet ($64 \times 64$)} & \textbf{ImageNet ($256 \times 256$)} \\ \hline

% Image resolution & $32 \times 32$ & $64 \times 64$ & $64 \times 64$ & $256 \times 256$ \\
Image size & 32 & 64 & 64 & 32 \\
Patch size & 2 & 4 & 4 & 2 \\
Input channels & 3 & 3 & 3 & 4 \\
Embedding dimension & 512 & 512 & 768 & 1,024 \\
Number of layers$^*$ & 13 & 13 & 17 & 21 \\
Number of heads & 8 & 8 & 12 & 16 \\
Number of classes & - & - & 1,000 & 1,000 \\
Latent space diffusion & No & No & No & Yes

\end{tabular}
\end{table}

\begin{table}[h]
\caption{\textbf{Training hyperparameters.} Training hyperparameters for the baseline, AdaDiff, and DuoDiff. \\} % Unless otherwise specified, the values are the same for all models.\\}

\label{tab:hp-training}
\centering

\begin{tabular}{l|l}
% \hline
 \textbf{Parameter} & \textbf{Training value} \\ \hline

% Image resolution & $32 \times 32$ & $64 \times 64$ & $64 \times 64$ & $256 \times 256$ \\
Training iterations &  \\
\quad U-ViT (CIFAR10 and CelebA) & 500,000 \\
\quad U-ViT (ImageNet) & 300,000 \\
\quad AdaDiff (output heads and UEMs) & 100,000 \\
Batch size & 128 \\
Optimizer & AdamW \\
Learning rate & 2e-4 \\
Weight decay & 3e-2 \\
$\beta_1$ & 0.99 \\
$\beta_2$ & 0.999 \\
Warmup steps & 1,500 \\
\end{tabular}
\end{table}

\end{document}